\documentclass[10pt, journal, twoside, preprint]{IEEEtran}

\IEEEoverridecommandlockouts 

\usepackage{cite}
\usepackage{amsmath,amssymb,amsfonts}
\usepackage{algorithmic}
\usepackage{graphicx}
\usepackage{subcaption}
\usepackage{booktabs}
\usepackage{cleveref}
\usepackage{threeparttable}
\usepackage{enumitem}
\usepackage{textcomp}
\usepackage{soul}
\usepackage{xcolor}
\usepackage{color,soul}
\usepackage{multirow}
\usepackage{svg}
    
\begin{document}

\title{Exo-Muscle: A Semi-Rigid Assistive Device  for the Knee}
\author{Yifang Zhang$^{1}$, Arash Ajoudani$^{2}$, and Nikos G. Tsagarakis$^{1}$
\thanks{This work was supported by the European Union’s Horizon 2020 research and innovation programme under Grant Agreement No. 871237 (SOPHIA).}
\thanks{$^{1}$Yifang Zhang and Nikos G. Tsagarakis are with the Humanoids and Human Centered Mechatronics (HHCM) research line of Istituto Italiano Di Tecnologia (IIT), Via S. Quirico, 19d, 16163 Genova, Italy 
        {\tt\footnotesize yifang.zhang@iit.it; nikos.tsagarakis@iit.it.}}
\thanks{$^{2}$Arash Ajoudani is with the  Human-Robot Interfaces and Physical Interaction (HRI$_{2}$) research line of Istituto Italiano Di Tecnologia (IIT), Via S. Quirico, 19d, 16163 Genova, Italy 
        {\tt\footnotesize  arash.ajoudani@iit.it}}
\thanks{Digital Object Identifier (DOI): 10.1109/LRA.2021.3100609.}
}

\markboth{Preprint Version. Accepted by RAL on Oct, 2021}
{Zhang \MakeLowercase{\textit{et al.}}: Exo-Muscle: A Semi-Rigid Assistive Device  for the Knee}

\maketitle

\begin{abstract}
In this work, we introduce the principle, design and mechatronics of Exo-Muscle, a novel assistive device for the knee joint. Different from the existing systems based on rigid exoskeleton structures or soft-tendon driven approaches, the proposed device leverages a new semi-rigid principle that explores the benefits of both rigid and soft systems. The use of a novel semi-rigid chain mechanism around the knee joint eliminates the presence of misalignment between the device and the knee joint center of rotation,  while at the same time, it forms a well-defined route for the tendon. This results in more deterministic load compensation functionality compared to the fully soft systems. The proposed device can provide up to $38Nm$ assistive torque to the knee joint. In the experiment section, the device was successfully validated through a series of experiments demonstrating the capacity of the device to provide the target assistive functionality in the knee joint.
\end{abstract}


\section{Introduction}

\IEEEPARstart{T}{he} development of wearable exoskeleton devices has been one of the main research areas in robotics during the past years, targeting to serve several applications from human rehabilitation to providing physical assistance for workers in heavy tasks. Various devices have been realized based on different arrangements, from multi-dof exoskeleton systems to modules targeting to support specific joints of the human body. 
Some typical examples of such devices are the one degree of freedom exoskeleton developed in \cite{385730} and \cite{7404039} to provide assistance to the knee and ankle joints, the hydraulic lower extremity exoskeleton BLEEX \cite{BLEEX}, the quasi-passive exoskeleton \cite{7001126} for load-carrying augmentation, and the active leg exoskeleton for gait training of stroke patients \cite{4663875}. The devices make use of rigid exoskeletons mounted along the human body joints of interest, assuming however simplified human joint models, e.g., they do not take into account the translation of the center of rotation of human joints due to their biological geometry with the complex articulating surface \cite{KneeMri}. This may induce misalignment issues between the biological and exoskeleton joints resulting in the generation of undesired constraints and forces and eventually discomfort.
\begin{figure}[t]
	\centering
	\includegraphics[width=0.9\linewidth]{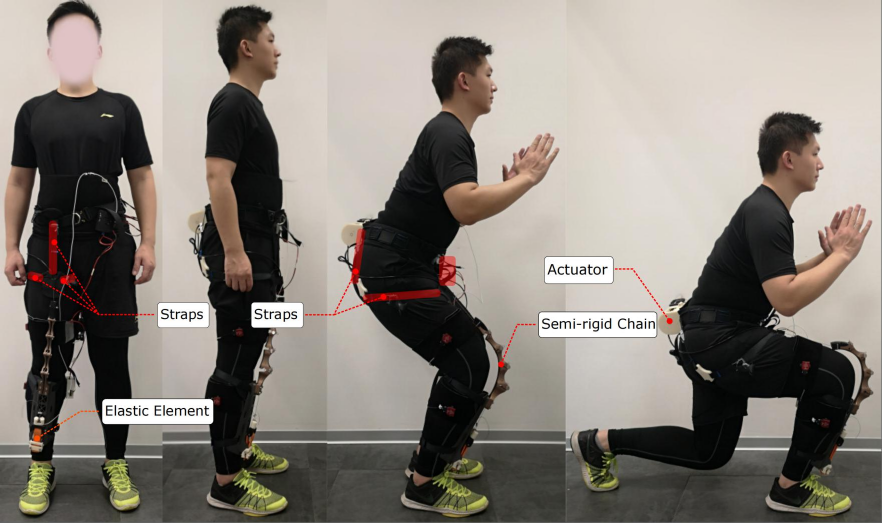}
	\caption{The Exo-Muscle device worn by a human subject in different knee poses. The main components of the device are indicated.}
	\label{Overall_Concept}
\end{figure}
Several solutions have been proposed to address this issue. Based on their principle, they can be categorized as rigid or soft systems. Rigid devices like the It-knee \cite{7759140} and ASSISON-KNEE \cite{6696472} explore redundant DOFs to compensate for the knee joint movement in the sagittal plane and deliver pure torque to the knee joint. The GEMS device \cite{8736810} and bio-knee joint \cite{5979761} achieve alignment with the instantaneous knee center of rotation, which uses cam and slider mechanisms. Concerning the soft actuation devices, the pneumatic actuated exo-suit was explored for gait assistant \cite{6631046} and knee joint rehabilitation \cite{6c23299}. Two different cable-driven exo-suits in \cite{6907024} were developed to assist the ankle plantar flexion and hip flexion/extension. Both rigid and soft actuation principles have their advantages and drawbacks. While the rigid solutions permit the generation of deterministic assistive torques that are more accurate and controllable, their approach to eliminating the misalignment issues by incorporating additional degrees of freedom results in more complex and heavier systems compromising ergonomics. Soft solutions can address the misalignment issue thanks to their inherent elasticity, but they usually require anchoring on the human body or cloth, which may generate undesired load directly on the human joints resulting in discomfort. Some solutions have been proposed for solving this issue by extending the support structure to create extra clearance. However, these designs can not still prevent contact between the tendons and the human knee joint \cite{10.3389/fnbot.2017.00057}. Besides, the moment arm generated by these devices around the human joint may vary during the joint motion, which makes it more challenging to provide a deterministic level of assistive torque.

To address the above limitations, we propose a new concept of a semi-rigid device for the knee joint (as shown in Fig.~\ref{Overall_Concept}) leveraging on the following novel features:  
\begin{itemize}
    \item  Use of a semi-rigid chain, misalignment and interference-free mechanism to route the transmission tendon of the device around the knee joint. 
    \item  Generation of fully deterministic assistive torque,  thanks to the transformation of the chain mechanism to a rigid structure that permits to route the tendon transmission along a certain path around the knee joint. This is not possible with a fully soft-tendon driven approach as the routing path may vary during the operation.
    \item Employment of a novel strap routing solution that distributes the load on the drive train to load insensitive areas of the human limb and improve the comfort and ergonomics of the device.
\end{itemize}
 The proposed solution addresses the limitations of both rigid and soft solutions while delivering their advantages in terms of deterministic assistive functionality and adaptation to the variations of the human knee joint center of rotation. In addition to the main novel features, a series elastic element in the drive train enables impact absorption and passive load compensation similar to other series elastic actuation driven devices \cite{doi:10.1177/0278364906063829,7281176,doi:10.1177/0278364915598388}. The paper is structured as follows. Section II introduces the design features of the Exo-Muscle device. Section III describes the control of the system. Section IV introduces the experiments performed and the results obtained. Finally, Section V draws the conclusion and discusses future activities.

\section{Exo-Muscle Design Principle and Implementation}
\begin{figure}[htbp]
	\centering
	\includegraphics[width=0.85\linewidth]{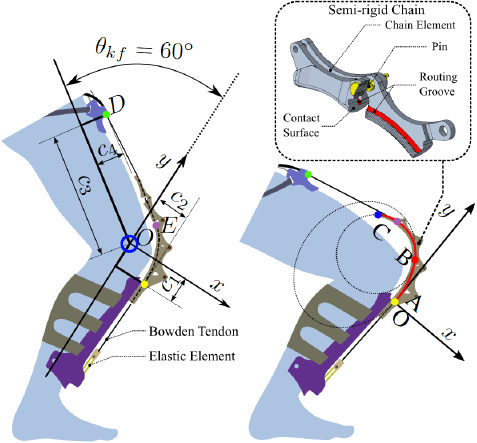}
	\caption{The semi-rigid chain concept at two knee flexion angles (left at $60^\circ$ and right at $100^\circ$).}
	\label{Chain_Working}
\end{figure}

The design of the Exo-Muscle device was driven by the following requirements.
\begin{itemize}
    \item The device should not generate any undesired forces and constraints caused by misalignment due to the translation of the human knee joint rotation axis.
    \item It should deliver deterministic load compensation functionality with low sensitivity to potential attachment inaccuracies, requiring no particular calibration and attention during the donning phase.
    \item It should demonstrate adaptable and soft interaction behaviour while providing assistive functionality.
    \item The structural complexity and weight of the device shall be minimized to reduce wearability fatigue.
\end{itemize}
Addressing all the above requirements is not possible by considering either a fully rigid or soft design approach. This motivated us to explore a semi-rigid principle. As shown in Fig.~\ref{Overall_Concept} and Fig.~\ref{Chain_Working}, the design features include a semi-rigid chain, which can transfer between rigid and flexible forms depending on the knee flexion angle and create a certain routing path for the tendon; a series elastic element, which enables impact absorption and passive load compensation, and a novel strap system for improving the comfort-of-use. The details of these features are presented in the following subsections.
\subsection{Semi-rigid Chain Mechanism}
An overview of the Exo-Muscle device concept is introduced in Fig.~\ref{Chain_Working}. One of the novel features of the Exo-Muscle is the guiding structure for its tendon system, which makes use of a semi-rigid chain mechanism in order to combine the advantages of both rigid and soft assistive device approaches.  Starting from the Bowden tendon termination on the thigh (indicated by the green point), the tendon extends from the attachment and cross over the routing groove formed by the semi-rigid chain, finally connected to the elastic element in the calf attachment. The guiding route is indicated as the red line in Fig.~\ref{Chain_Working}, and it can be described by
\begin{equation}\label{eq1}
f\left(x\right)=
\begin{cases}
b\sqrt{1 - \frac{\left(x + x_e\right)^2}{a^2}} - y_e & -0.034 \leq x  \leq 0\\
\sqrt{R^2 - \left(x + x_c\right)^2} - y_c & -0.171  \leq x  < -0.034
\end{cases}
\end{equation}
where $f\left(x\right)$ defines the guiding route in the frame shown in the right-hand side of Fig.~\ref{Chain_Working}. $x_c$, $y_c$, $R$, $x_e$, $y_e$, $a$ and $b$ denote the constants of \eqref{eq1}. In this study, their values were adjusted base on the lower limb model shown in Fig.~\ref{Chain_Working} to ensure that the correlated semi-rigid chain can keep clearance with the human tissue in the operating range of the knee joint. Their values are given in Table ~\ref{tab2}. By adjusting these parameters and the design of chain elements, the device can be worn by persons of different body sizes.
\begin{table}[htbp]
	\caption{Parameters of profile equations}
	\begin{center}
		\begin{tabular}{|c|c|c|c|c|c|c|}
			\hline
			\multicolumn{3}{|c|}{\textbf{Circle}} & \multicolumn{4}{c|}{\textbf{Ellipse}} \\ \hline
			$x_c$    &   $y_c$    &    $R$   &  $x_e$   &  $y_e$   &  $a$   &  $b$   \\ \hline
			0.107   &   -0.076    &    0.09   &    0.125 & -0.015    & 0.125    &  0.165   \\ \hline
			\multicolumn{7}{l}{\textit{Note:}Values are rounded to 3 significant decimal digits.}
		\end{tabular}
		\label{tab2}
	\end{center}
\end{table}
As shown in Fig.~\ref{Chain_Working}, there are three important points $A$, $B$ and $C$, on the guiding route. The yellow point $A$ indicates the start point of the chain; the red point $B$ is the transition point where the profile of the chain changes from the ellipse to the circle; the blue point $C$ is the endpoint of the semi-rigid chain. The coordinates of these points are $A\left(0, 0.015\right)$, $B\left(-0.034, 0.128\right)$ and $C\left(-0.171, 0.139\right)$. This guiding profile ensures that the tendon can cross over the human knee without any contact with the knee, even in large knee flexion angle conditions. 
Finally, the designed continuous guiding route is represented by a modular designed semi-rigid chain. It consists of five basic elements, which are illustrated in the top right of Fig.~\ref{Chain_Working}. The routing groove of each element represents a segment of the guiding route (1). As shown in the left-hand side of Fig.~\ref{Chain_Working}, when the device is attached to a user's lower limb, the tendon from the thigh attachment tangentially engages with the designed guiding profile at the purple point E. The elements between points $A$ and $E$ reach their mechanical limit defined by the contact surfaces and discretely compose a part of the continuous guiding profile (1). At the same time, elements after the point $E$ flexibly align on the tendon between point $D$ and $E$. With the increase of knee flexion angle (KFA), as shown in the right-hand side of Fig.~\ref{Chain_Working}, more elements will reach their mechanical limit and extend the guiding route until the designed guiding route (1) is fully developed.

\begin{figure}[htbp]
\centering
        \subfloat[]{\includegraphics[width=0.36\linewidth]{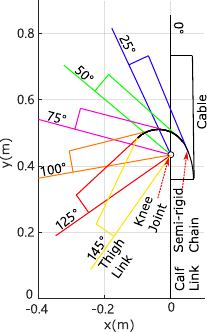}\label{Chain_Simulation:a}}
        \hfill
        \subfloat[]{\includegraphics[width=0.50\linewidth]{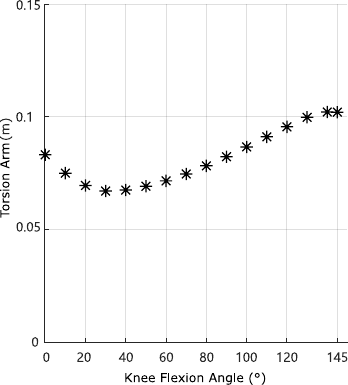}\label{Chain_Simulation:b}}
        \caption{ (a) Visualization of simulation in MATLAB. (b) The change of torsion arm during the increase of knee flexion angle.}
        \label{Chain_Simulation}
\end{figure}

Simulation studies were performed in MATLAB to evaluate the moment arm formed by this semi-rigid chain structure while the KFA $\theta_{kf}$ is varying from $0^\circ$ to $145^\circ$. As shown in the left side of Fig.~\ref{Chain_Working}, the frame applied in the simulation located at the knee joint, and the joint was simplified as a re-volute joint. The y-axis of the frame aligns with the calf's longitudinal axis, while the x-axis is parallel with the sagittal plane and perpendicular to the y axis. The representation of semi-rigid chain model \eqref{eq1} in the frame located at the knee joint is described by $f_k\left(x\right)$
\begin{equation}\label{eq2}
f_k\left(x\right) = f\left(x - c_2\right) - c_1
\end{equation}
where $c_1$ and $c_2$ denote the distance between the start point $A$ of the semi-rigid chain and the knee rotation center in $y$ and $x$ directions, respectively. The green point $D$ indicates the Bowden tendon housing end on the thigh. Its position $P_D$ with respect to the knee joint frame depends on the knee flexion angle $\theta_{kf}$.
\begin{equation}\label{eq3}
P_D = 
\begin{bmatrix}
\cos{\theta_{kf}}&-\sin{\theta_{kf}}\\
\sin{\theta_{kf}}&\cos{\theta_{kf}}
\end{bmatrix}
\begin{bmatrix}
c_4\\
c_3
\end{bmatrix}
\end{equation}
where $c_3$ and $c_4$ denote the distance between point $D$ and knee rotation center in $y$ and $x$ directions when $\theta_{kf}$ is $0^\circ$ respectively. As the tendon is tangent with the routing groove of the semi-rigid chain at the purple dot $E$, it should then satisfy the following equation:
\begin{equation}\label{eq4}
f_{k}^{'}\left(P_{Ex}\right) = \frac{f_{k}\left(P_{Ex}\right) - P_{Dy}}{P_{Ex} - P_{Dx}}
\end{equation}
where the $f_{k}^{'}\left(P_{Ex}\right)$ indicates the first-order derivative of \eqref{eq2} at the point $E$. The $P_{Dx}$, $P_{Ex}$ and $P_{Dy}$ denote the $x$ and $y$ components of points $D$ and $E$ with respect to the knee joint frame. \eqref{eq4} is solved with $vpasolve\left(\right)$ function in MATLAB to get the $P_{Ex}$ at different knee flexion angles. The moment arm $l_{a}$ of the device is then given by
\begin{equation}\label{eq5}
l_{a} = 
\frac{\begin{Vmatrix}\boldsymbol{DE} \times \boldsymbol{DO}\end{Vmatrix}}
{\begin{Vmatrix}\boldsymbol{DE}\end{Vmatrix}}
\end{equation}
where the $\boldsymbol{DE}$, $\boldsymbol{DO}$ and $\boldsymbol{DE}$ denote the corresponding vector in the knee joint frame. The simulation model is visualized in Fig.~\ref{Chain_Simulation:a}. Fig.~\ref{Chain_Simulation:b} shows the torsion arm profile, which demonstrates a slight decrease trend between $0^\circ$ to $30^\circ$ KFA, while it increases for $KFA>30^\circ$. Overall, the designed routing profile ensures that the cable can cross over the human knee joint without applying force on the human tissue around the knee. Simultaneously, it creates a predefined and deterministic guiding route for the tendon to generate a desired torque profile for the human knee. The torsion arm profile result was fitted to \eqref{eq6}
\begin{multline}\label{eq6}
L_{a}\left(\theta_{kf}\right) = -2.271e^{-12} \cdot {\theta_{kf}}^{5} + 1.096e^{-9} \cdot {\theta_{kf}}^{4}\\
- 2.462e^{-7} \cdot {\theta_{kf}}^{3} + 2.87e^{-5} \cdot {\theta_{kf}}^{2}\\
- 1.012e^{-3} \cdot \theta_{kf} + 0.074
\end{multline}
where the $L_{a}\left(\theta_{kf}\right)$ denotes the moment arm at knee flexion angle $\theta_{kf}$. As solving \eqref{eq4} in the control loop may cause latency, \eqref{eq6} was used in the controller as described in the next sections.

\subsection{Elastic Element Selection}

\begin{figure}[htbp]
	\centering
	\subfloat[]{\includegraphics[width=0.40\linewidth]{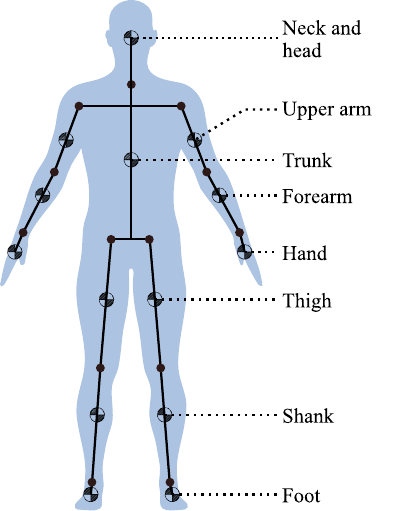}\label{LoadSimulation:a}}
	\hfill
	\subfloat[]{\includegraphics[width=0.41\linewidth]{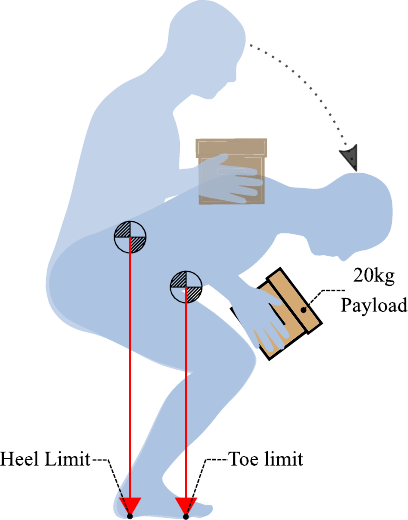}\label{LoadSimulation:b}}
	\caption{ (a) Human body link mass model. (b) Indication of a posture used in the simulation (ADA: $-30^\circ$, KFA: $90^\circ$, HFA is regulated to move the COP from heel limit to toe limit.).}
	\label{LoadSimulation}
\end{figure}

For the selection of the elastic element in series with the tendon transmission, a study was performed on a simplified human joint-link mass model in order to estimate the torque generated at the human knee joint. Fig.~\ref{LoadSimulation:a} introduces the human model and COM of each segment used in the simulation. The joints are located in the human's frontal plane, and the DoFs for each joint are reduced to one DoF rotation in the direction of a sagittal plane. The corresponding length, mass and COM position values of the body segments were defined based on \cite{DELEVA19961223}. 
According to \cite{BBHM}, the motion ranges of ankle dorsiflexion, knee flexion and hip flexion were set to $0^{\circ}$ to $-40^{\circ}$, $0^{\circ}$ to $145^{\circ}$ and $-15^{\circ}$ to $125^{\circ}$ respectively. Fig.~\ref{LoadSimulation:b} shows the posture in the simulation for ankle dorsiflexion angle (ADA) at $-30^{\circ}$ and KFA at $90^{\circ}$. A number of simulations were performed in the workspace of the ankle and knee joints in which the upper body posture remained constant, and a $20kg$ payload was added at hand. The hip flexion angle (HFA) was varying from $125^\circ$ to $-15^\circ$ to move the center of pressure (COP) from the heel to the toe limit. A URDF file of the simplified human joint-link model and the dynamic computations library Pinocchio \cite{pinocchioweb} were used to estimate the output torque generated at the knee joint. The estimated torque was divided by the moment arm of the device at a corresponding KFA to compute the required tension force for the tendon. A maximum required tendon tension force of $323.9N$ was computed at the pose: ADA $-30^\circ$, KFA $50^\circ$, HFA $14^\circ$.

\begin{figure}[htbp]
\centering
        \includegraphics[width=0.9\linewidth]{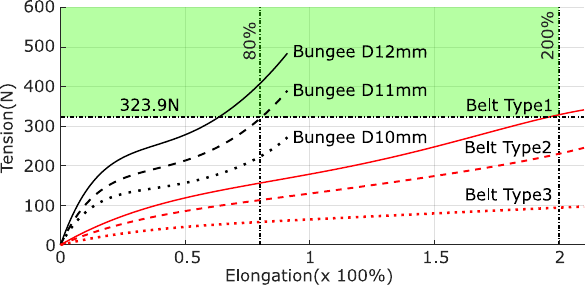}
        \caption{Tensile test results for three types of rubber belts (type1: $27.5 \times 4.15mm$, type2: $24 \times 4mm$ and type3: $13 \times 4.5mm$ respectively) and tensile curves of the bungee cords provided by Shandow Technic.}
        \label{Rubber_Test}
\end{figure}

For the selection of the elastic element, two types of elastic elements, a rubber belt from DOMYOS and a bungee cord from Shandow Technic, were evaluated.
Fig.~\ref{Rubber_Test} shows the force curves of the bungee and the rubber belt. The recommended elongation range for bungee and rubber belt are $0$ to $80\%$ and $0$ to $200\%$ respectively. The viscoelastic effects of the materials are neglected in this study since the human motions in the experiments are low-speed, and the proposed force feedback controller can compensate for deviations caused by these effects. The elastic elements that can enter the green zone in Fig.~\ref{Rubber_Test} within their working range can satisfy our load capacity requirement. Thanks to its larger elongation range, the initial length of the belt is shorter than that of the bungee. Furthermore, the belt section shape and the simpler end fitting permit the more compact integration of the belt elastic element $Type-1$ in the calf attachment. For these reasons, the belt elastic element was selected. With this configuration, the device can provide up to $38Nm$ assistive torque for the knee joint when the KFA is $145^\circ$.

As the tension is nonlinear with respect to the elongation, its elastic model was fitted to \eqref{eq8}
\begin{multline}\label{eq8}
Y\left(F\right) = 2.803e^{-12} \cdot F^{5} - 2.291e^{-9} \cdot F^{4} + 5.882e^{-7} \cdot F^{3}\\
- 3.83e^{-5} \cdot F^{2} + 3.674e^{-3} \cdot F
\end{multline}
where $Y\left(F\right)$ indicates the elongation of the rubber belt at tension force $F$. \eqref{eq8} was used by the control system and for the generation of the assistive torque around the knee.
\begin{figure}[htbp]
	\centering
	\subfloat[]{\includegraphics[width=0.42\linewidth]{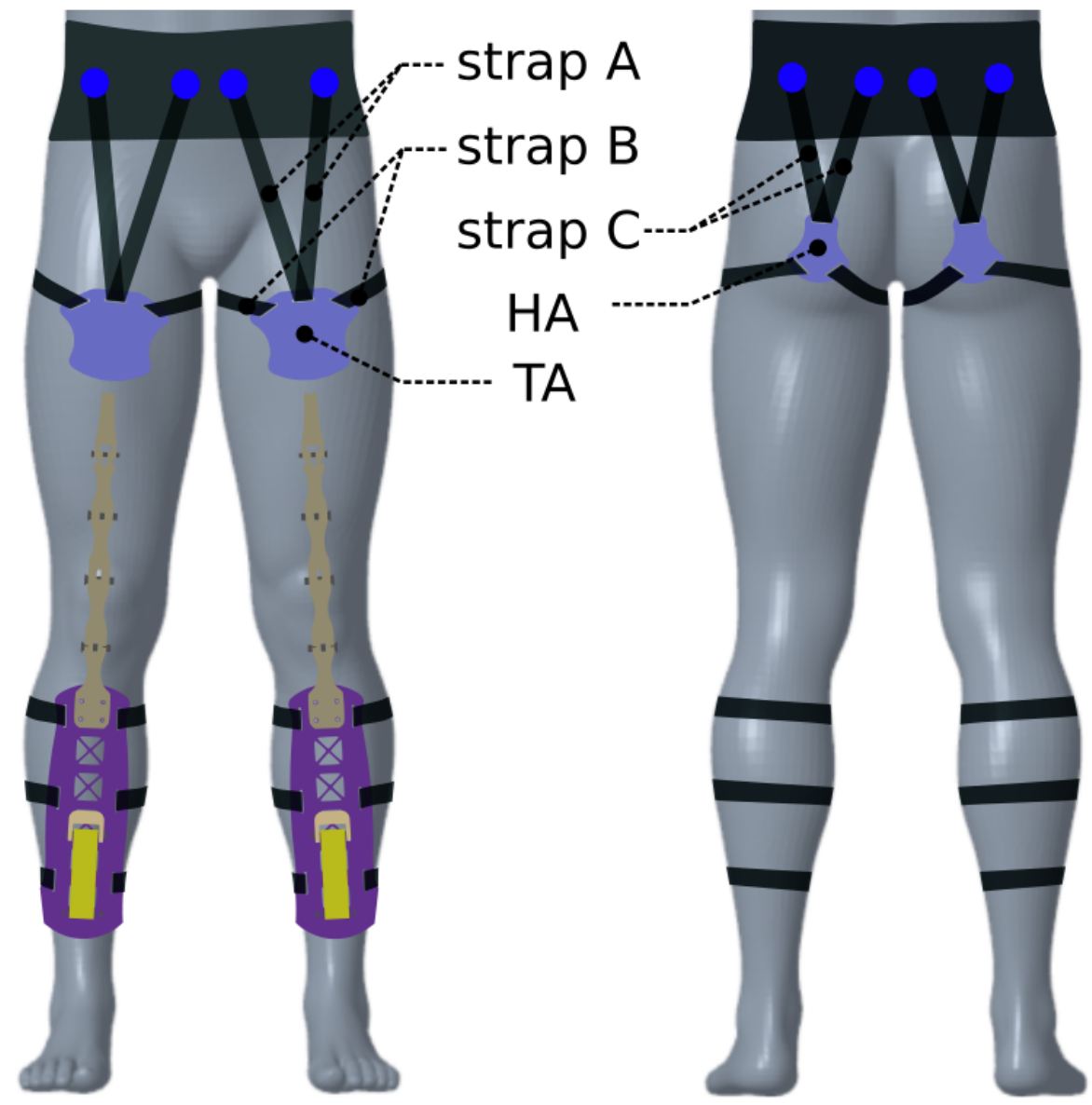}\label{Body_Attachment:a}}
	\hfill
	\subfloat[]{\includegraphics[width=0.44\linewidth]{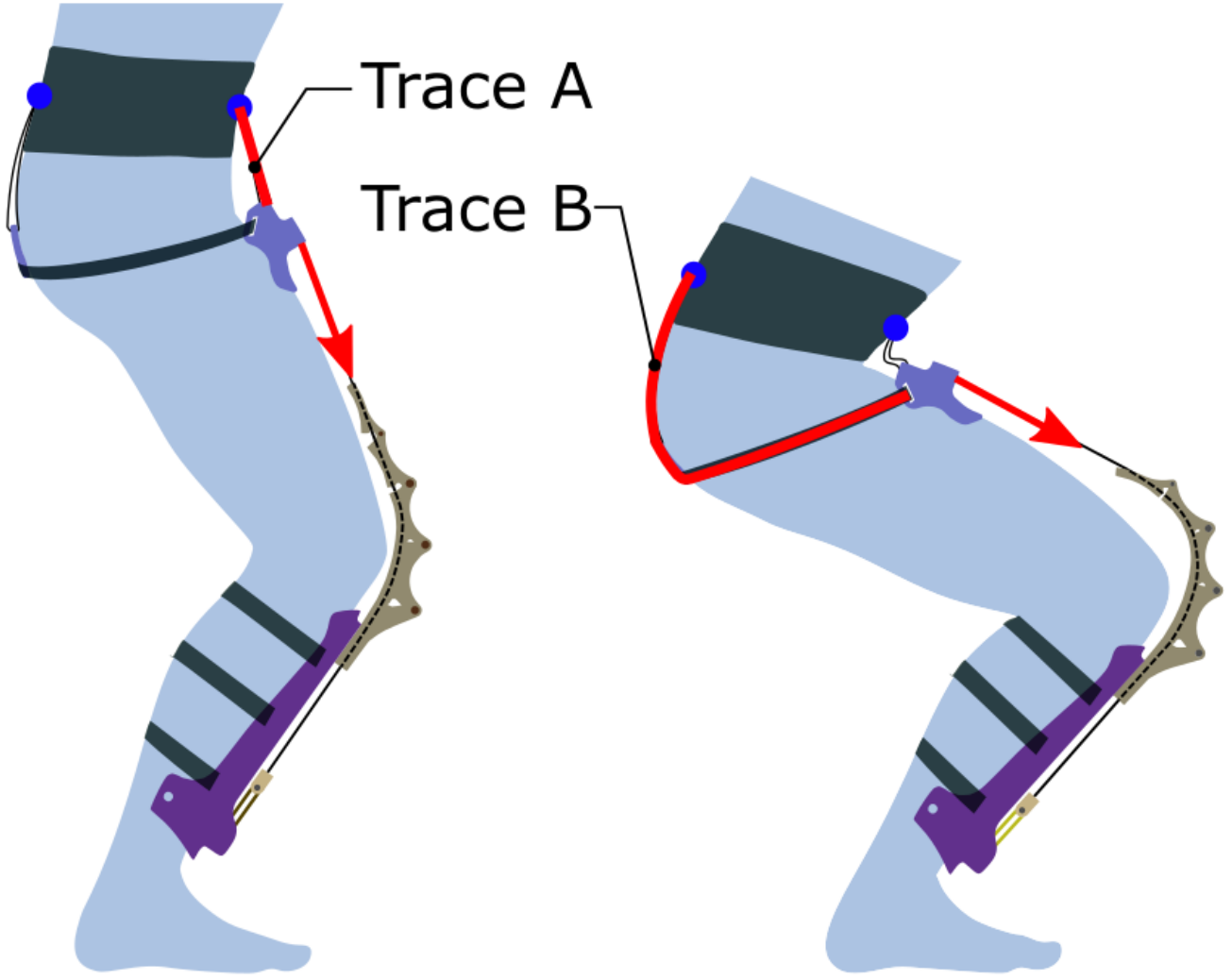}\label{Body_Attachment:b}}
	\caption{ (a) Indication of strap routing. (b) Load traces in different posture (Left: ADA $-20^\circ$, KFA $50^\circ$, HFA $-40^\circ$. Right: ADA $-30^\circ$, KFA $90^\circ$, HFA $-70^\circ$).}
	\label{Body_Attachment}
\end{figure}
\subsection{Strap routing Design}

In our system, the maximum tension force on the tendon is $323.9N$. Directly anchoring this high tension load near the knee joint or thigh would cause discomfort near the joint and unrecoverable sliding of the anchor point. To tackle this, we explore a novel webbing strap routing method that can directly anchor the end of straps to the waist to obtain a stable attachment and distribute the load to the upper thigh, waist and back of the hip joint, improving the user comfort during operation. As it can be seen in fig.~\ref{Body_Attachment:a}, the proposed routing of the supporting straps involves an attachment on the front side of the thigh that provides one anchor point for the Bowden tendon transmission. This attachment point is connected with three webbing straps. Strap A is directly linked to the waist belt on the front side. Strap B bypass the thigh's inner side and outer side and connect to the hip attachment at the backside of the hip. Finally, strap C links the hip attachment and the backside of the waist belt. Strap A forms the load trace A, while the straps B and C form the load trace B. Depending on the depth of the squat motion, the load distribution switches between these two load traces. As can be observed from fig.~\ref{Body_Attachment:b}, at the beginning of squat, the strap trace A will deliver the load to the waist belt while the trace B is loose. With the increase of the squat depth, the mounting point of trace A will move toward the knee joint while the length of the required straps of trace B is increasing. Thus, trace A gradually becomes loose, and the tension load will deliver to the backside of the waist belt via trace B that has the additional benefit of generating a positive torque around the hip supporting this joint during the deep squat.

\section{Control}

\begin{figure*}[htbp]
	\centering
	\includegraphics[width=1\linewidth]{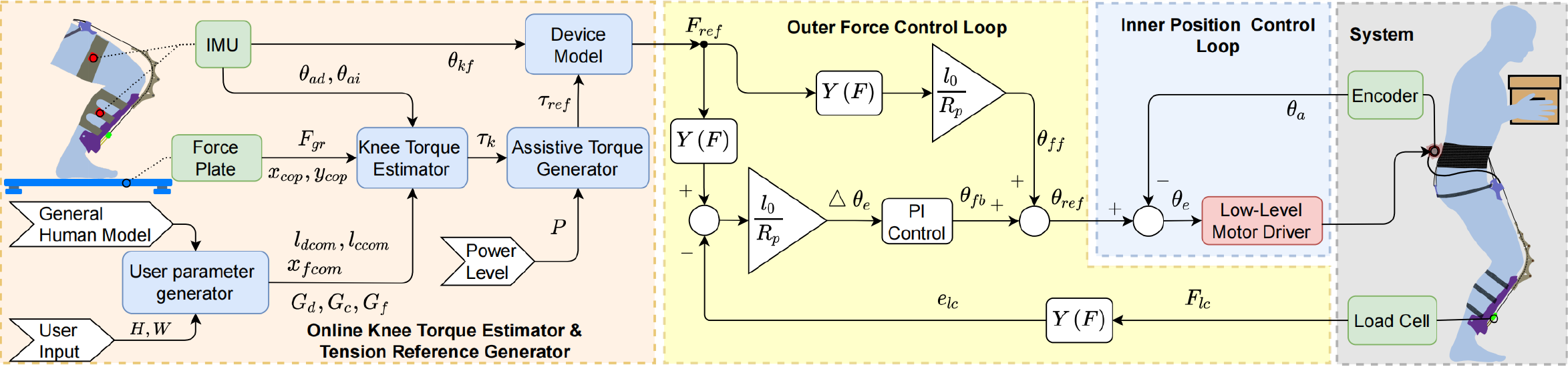}
	\caption{The control schematic depicting the online knee torque estimation and the tendon force regulation components.}
	\label{Controller}
\end{figure*}

This section introduces the control system used to regulate the assistive torque generated by the Exo-Muscle device. Fig.~\ref{Controller} presents an overview of the Exo-Muscle control system indicating the sensory and control components and their interaction with the human model and motion inputs. The white arrows indicate the user interface for a defined user's body parameter and desired assistance level.

\subsection{Sensing and Instruments}
For the control of the device, two IMU sensors (VN-100T, VectorNav) were attached to the calf and thigh of the human subject for monitoring the pose of the calf and thigh, respectively, as illustrated Fig.~\ref{Controller}. The data from the IMU was post-processed to obtain the ADA $\theta_{ad}$, inversion angle (AIA) $\theta_{ai}$ and KFA $\theta_{kf}$. A load cell (8417-6001, burster) was integrated between the Bowden tendon and the rubber belt. The measured tension force is noted as $F_{lc}$. A force plate was used to measure the ground reaction force $F_{gr}$ and obtain the location of COP, $x_{cop}$ and $y_{cop}$ of the human subject while standing on the force plates and varying the lower limb posture. The actuation of the Exo-Muscle prototype makes use of a highly integrated motorized actuator (TREE Lime, \cite{8630605}). The actuator is equipped with position (motor and link side encoders) and torque sensing. 
An embedded PC (COM Express) running a real-time system Xenomai 3 and an EtherCAT master module was used to enable real-time control and communication through EtherCAT with the Exo-Muscle actuator and sensors. The communication bandwidth with the Exo-Muscle actuator and sensors was set at 1kHz.

\subsection{Online Knee Torque Estimation}
For computing the reference tension force for the controller, the human knee torque is firstly estimated by considering the ground reaction force measured by the force plate and the posture of the human ankle, foot and calf derived by the IMU measurements \cite{8009423}. 
The knee joint torque can be estimated from the torque generated by the ground reaction force by subtracting the torques generated by the lower leg segments (calf and foot) as well as torque produced due to the mass of the device components mounted on the calf.
\begin{equation}\label{eq9}
\tau_{k} = \tau_{gr} -\left(\tau_{d}+\tau_{c}+\tau_{f}\right)
\end{equation}
where $\tau_{k}$ denotes the knee joint torque. The $\tau_{c}$, $\tau_{f}$ and $\tau_{d}$ are the torque at the knee joint induced by the gravity force of the calf, foot and device. The  $\tau_{gr}$ denotes torque generated by the ground reaction force. These torque components can be calculated with \eqref{eq10}, \eqref{eq11}, \eqref{eq12} and \eqref{eq13}.
\begin{align} 
\tau_{d} = - G_{d}\cos\theta_{ai}  \cdot  l_{dcom} \sin\theta_{ad} \label{eq10} \\ 
\tau_{c} = - G_{c}\cos\theta_{ai}  \cdot  l_{ccom} \sin\theta_{ad} \label{eq11} \\ 
\tau_{f} = - G_{f}\cos\theta_{ai}  \cdot  \left(l_c \sin\theta_{ad} - y_{fcom}\right) \label{eq12} \\ 
\tau_{gr} = F_{rgr}\cos\theta_{ai}  \cdot  \left(l_c \sin\theta_{ad} - y_{cop}\right) \label{eq13}
\end{align}
where $G_{d}$, $G_{c}$ and $G_{f}$ denote the gravity force of the device, calf and foot. The $l_{dcom}$ and $l_{ccom}$ are the distance for the CoM of the device and calf to the knee joint axis. Their CoM are assumed to be located on the link model of the calf. The $l_c$ is the length of the calf link. The $y_{fcom}$ is the $y$ coordinate of the foot's CoM in the world frame. The ground reaction force on the right foot $F_{rgr}$ can be computed by
\begin{equation}\label{eq14}
F_{rgr} = F_{gr} \cdot \frac{2x_{cop}+l_fs}{2l_fs}
\end{equation}
The $l_fs$ denotes the distance between the two ankle joints. When $x_{cop} = \frac{l_fs}{2}$ the right foot will experience the entire weight of the human subject resulting in $F_{rgr} = F_{gr}$.

\subsection{Controller design}
The Exo-Muscle assistive torque regulation is achieved through the control of the tendon force. As it can be seen in Fig.~\ref{Controller}, an outer force controller receives the reference tendon force $F_{ref}$ computed by
\begin{equation}\label{eq15}
F_{ref} = \frac{\tau_{k} \cdot P}{L_a\left( \theta_{kf} \right)}
\end{equation}
where, $\tau_{k}$ is the knee torque, estimated by \eqref{eq14}, P denotes the user-defined compensation percentage  level ($0$ to $100\%$) of the knee torque and $L_a\left( \theta_{kf} \right)$ is the torsion arm that is a function of the knee flexion angle, see \eqref{eq6}.

The reference tendon force is converted into the desired tendon displacement using the elastic element model of \eqref{eq8}.  Similarly, the tension force $F_{lc}$ measured by the load cell is used to compute the current tendon displacement by fitting $F_{lc}$ into the elastic element model of \eqref{eq8}. The current tendon displacement is then used as feedback for the outer control loop that employs a proportional-integral (PI) controller. This outer force control loop ensures the tracking of the required Bowden tendon tension force by generating the appropriate references for the inner position controller of the actuator, which eventually regulates the Bowden tendon displacement. 
The transmission efficiency was tested with the setup shown in \cite{Carlson1995EfficiencyOP}. The surface of the guiding route was finely machined and lubricated. As a result, the efficiency changed from $100\%$ to around $85\%$ when the cable turned from $0^\circ$ to $100^\circ$ on the guiding route. This is also the operation range of the device in the human experiments section. Thus, the device is operated in the high-efficiency range, and the friction between the Bowden tendon and semi-rigid chain is neglected in this work. The tension force is assumed to equal that measured by the load cell installed between the Bowden tendon and the elastic rubber belt. For improving the tracking dynamics of the tendon tension force regulator, a feedforward term is also added, see the yellow block in Fig.~\ref{Controller}.  The reference position of the actuator $\theta_{ref}$ is therefore given by
\begin{equation}\label{eq16}
\theta_{ref} = \theta_{ff} + \theta_{fb}
\end{equation}
where $\theta_{ff}$ and  $\theta_{fb}$ denote the output of feedforward and feedback branches. The feedforward term of the actuator position reference is computed by fitting the $F_{ref}$ into the elastic model of \eqref{eq8} as follows.
\begin{equation}\label{eq17}
\theta_{ff} = Y\left(F_{ref}\right) \cdot \frac{l_0}{R_p}
\end{equation}
where ${R_p}$ denotes the radius of the actuator output pulley where the Bowden tendon is attached, and ${l_0}$ represents the nominal zero dilated length of the rubber belt element. The feedback part of the actuator position reference is given by
\begin{equation}\label{eq18}
\theta_{fb} = K_{p} \cdot \bigtriangleup \theta_{e} + K_{i} \cdot \intop\nolimits_{0}^{t} \bigtriangleup \theta_{e}\left(t_d\right)d\end{equation}
\begin{equation}\label{eq19}
\bigtriangleup \theta_{e} = \left( Y\left(F_{ref}\right) - Y\left(F_{lc}\right)\right) \cdot \frac{l_0}{R_p}
\end{equation}
where $K_{p}$ and $K_{i}$ are the proportional and integral gains of the feedback loop. 
\begin{figure}[htbp]
	\centering
	\includegraphics[width=0.7\linewidth]{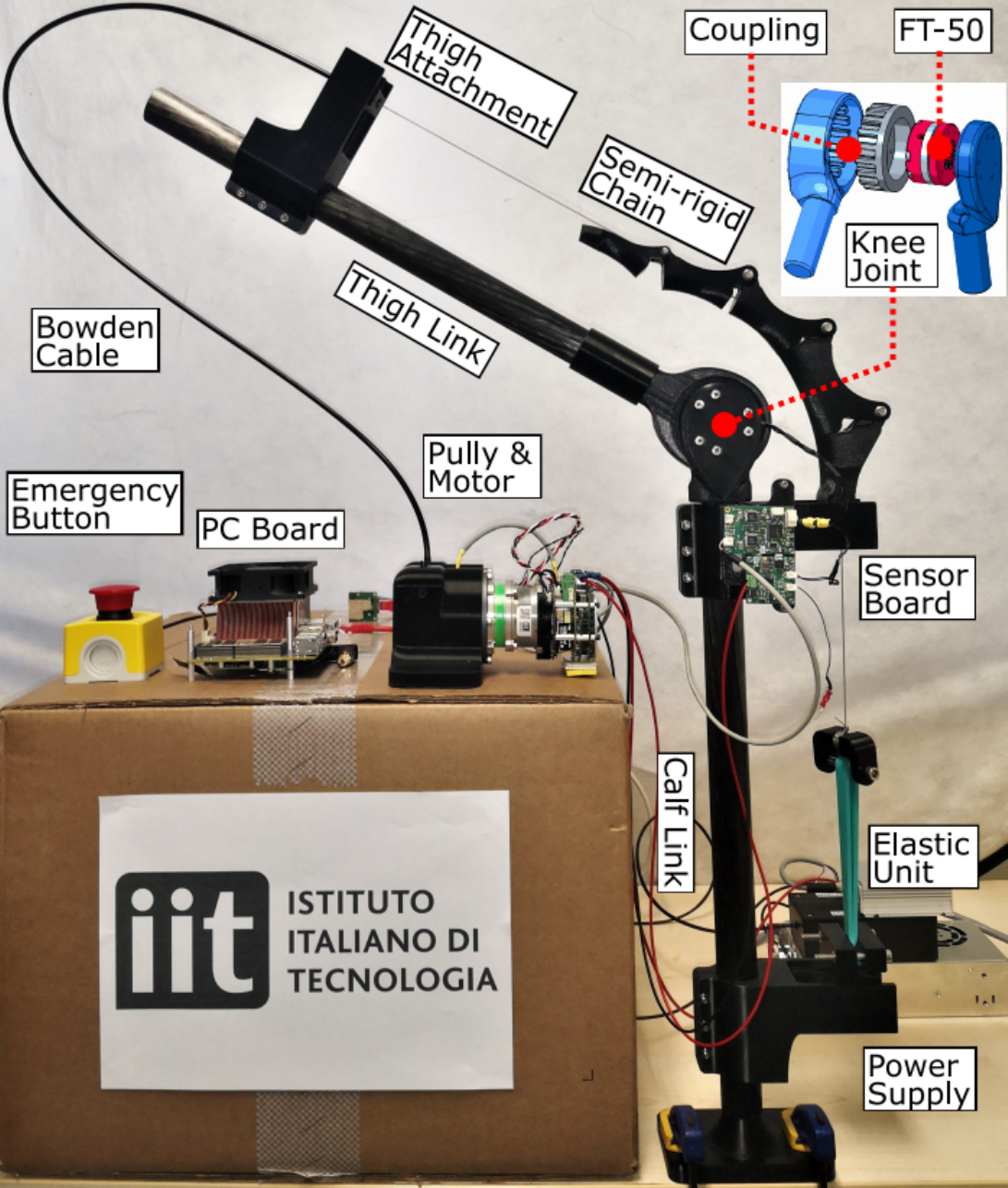}
	\caption{The mechanical knee joint testbed used to validate the working principle of the Exo-Muscle.}
	\label{Experimental_Setup}
\end{figure}
\begin{figure}[htbp]
\centering
    \subfloat[][]{\includegraphics[width=0.8\linewidth]{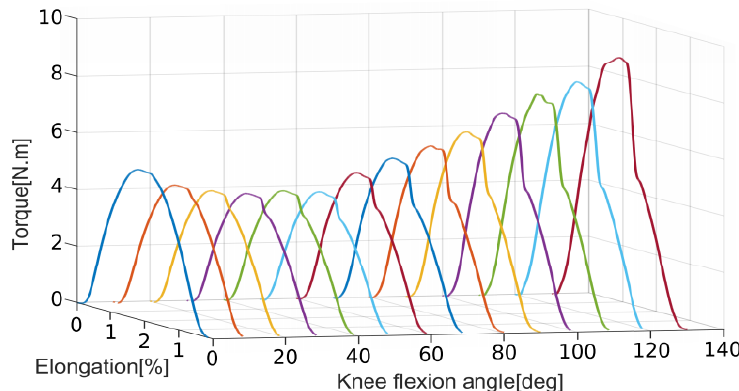}\label{Experimental_Dataplot:a}}
    \hfill
    \subfloat[][]{\includegraphics[width=0.9\linewidth]{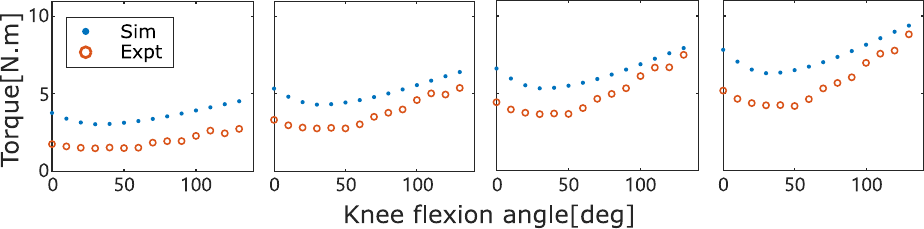}\label{Experimental_Dataplot:b}}
    \caption{Experimental results. (a) Torque-Elongation profiles under different knee flexion angle, over one load period. (b) Experimental and simulation results for different elongation conditions. (From left to right $50\%$,$100\%$,$150\%$ and $200\%$ elongation respectively).}
    \label{Experimental_Dataplot}
\end{figure}

\section{Experimental Validation}

For evaluating the performance of the developed Exo-Muscle assistive device, a number of experiments were carried out and discussed in the following subsections, together with the results obtained.  The Exo-Muscle device used in the experiments makes use of the TREE Lime actuator \cite{8630605} combined with an output tendon pulley of $45mm$ in diameter, enabling to generate up to $400N$ continuous tension force in the Bowden tendon. The total weight of the Exo-Muscle prototype is $2.5Kg$ distributed to the calf module ($0.85Kg$), the actuator module($1.3Kg$) and Bowden tendon transmission and attachments ($0.35Kg$). 

\subsection{Concept Validation}

\begin{figure*}[htbp]
	\centering
	\includegraphics[width=1\linewidth]{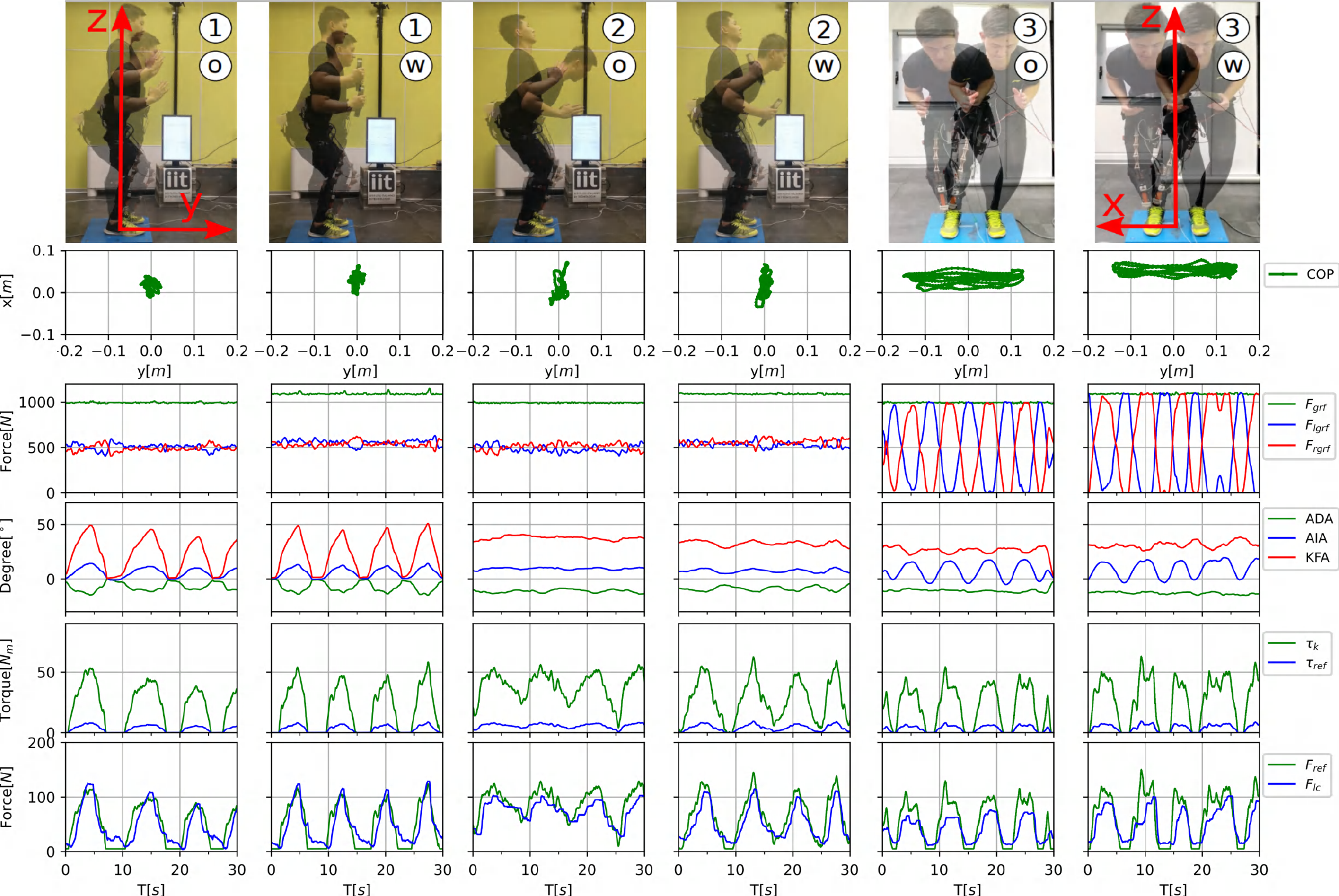}
	\caption{The experimental results from the  three motion cycles (from left to right) performed with the human subject. The mark "o" and "w" denote the "with" and "without" payload conditions).}
	\label{Human_experiments}
\end{figure*}
The first set of experiments targeted to evaluate the working principle of the device and its ability to generate assistive torque around a mechanical knee joint testbed, Fig.~\ref{Experimental_Setup}.

A spline coupling with an average distributed tooth is incorporated in the knee joint testbed, permitting to adjust the angle of the mechanical knee joint with $10^\circ$ increments from  $0^\circ$ to $130^\circ$. This enables adjustment of the knee joint testbed to 14 different knee configurations. A torque sensor, integrated into the knee-joint coupling, monitors the torque generated around the joint by the attached Exo-Muscle device. The semi-rigid chain, spline coupling and attachments of the mechanical knee joint testbed prototype were made of ABS plastic using 3D printing fabrication. Thus, the load capacity of the knee joint testbed is limited due to the structural strength of the ABS material. For this reason, the belt $Type-3$, which demonstrates the lowest stiffness in Fig.~\ref{Rubber_Test}, was used in the experiments with the knee joint testbed. After mounting the Exo-Muscle on the knee joint testbed, the actuator of the device was position controlled with a same trajectory to generate an extension on the elastic belt element until its maximum working elongation $200\%$. The elastic element used in the test has an initial length of $55mm$. The length change of the tendon is measured by the encoder of the Exo-Muscle actuator. The generated assisting torque around the kneed joint testbed is monitored by the torque sensor integrated into the knee joint testbed. The process was repeated at every angle configuration of the knee joint (14 data sets were collected in total).  The record data was processed with a low pass filter and plotted in the Fig.~\ref{Experimental_Dataplot:a}. The torque profile of the different KFA forms an asymmetrical \textbf{U} shape trend. Along the KFA axis, the torque slightly decreases with the increase of the KFA, until $30^\circ$,  while the torque rises with the further increase of the KFA. This trend matches with the torsion arm profile shown in Fig.~\ref{Chain_Simulation:b}, showing that the Exo-Muscle can generate the designed torque profile. In addition, Fig.~\ref{Experimental_Dataplot:b} shows the experimental (red circles) and simulation (blue dots) results for different elongations. The simulation results were obtained by multiplying the torsion arm shown in Fig, ~\ref{Chain_Simulation:b} with the tension force of the elastic belt at the corresponding elongation in Fig.~\ref{Rubber_Test}. Overall, the experimental trends match well the simulation profiles. The deviations observed are related to the low tolerances in the 3D printed spline coupling and the low structural stiffness and resulted in deformations of the ABS made thigh and calf links. Both issues caused the reduction of the elastic belt elongation and moment arm resulting in lower torque experimental profiles compared to the simulation ones.

\subsection{Human experiments}

In human experiments, the controller reference is the desired tension force $F_{ref}$ which is derived by the online knee torque estimator and tension reference generator module shown in Fig.~\ref{Controller}, and the higher stiffness belt $Type-1$, Fig.~\ref{Rubber_Test} was used in the Exo-muscle device. The device was worn by a human subject and assessed in three different motion cycles. Each motion cycle was performed under two conditions (with a payload of $10kg$ held by the user and without the payload). The whole experimental procedure was approved by the regional ethics committee of Liguria: studio IIT-HRII-SOPHIA – N. CER Liguria 554/2020). The first row of Fig.~\ref{Human_experiments}, indicate the poses and motions in each experiment. The plots in the second row show the traces of the COP. It can be observed that the location of COP is nearly constant for the standard stand to squat motion cycles 1. When keeping KFA and ADA fixed and adjusting the HFA, the COP motion along the y-direction can be observed in cycle 2. During motion cycle 3, the lateral shift of the COP between two feet in the x-direction is evident. By using the COP location, the distribution of the ground reaction force in each foot is computed from \eqref{eq14} and plotted in the third row. The ADA, AIA and KFA joint angle data measured by the IMU sensors attached on the right calf and right thigh are shown in the fourth row. The human knee joint torque was then estimated based on the equations presented in the Online Knee Torque Estimation section, considering the human model segment parameters scaled accordingly based on the height ($1.89m$) and weight ($100kg$) of the human subject. By comparing the $with$ and $without$ the payload of $10kg$ trials, a nearly equal value can be observed for cycle-1, as the payload is upright the knee joint during the execution of this motion cycle. In cycle-2, the maximum and minimum value of the estimated knee torque during the $with$ payload trial is higher and lower than those of the $without$ payload trial. This is because when the payload passes through the singular position of the knee joint, it counterbalances the torque generated by the human body when it moves to the front side, resulting in a decrease of the lowest torque value.
In contrast, it creates a higher knee torque when it moves to the backside. In cycle-3, the knee torque during the $with$ payload trial is generally lower than that of the $without$ payload condition as the payload stays in front of the knee joint during the execution of this cycle. These results indicate that the proposed method can effectively obtain the user's state and use it for deriving the knee torque. Then the torque reference (blue lines) was derived considering a preset compensation level of $15\%$. The green lines and blue lines in the sixth row indicate the desired force reference and measured tension force showing good tracking performance from the force controller.

\section{Conclusion \& Future Work}
A novel semi-rigid assistive device for the knee has been introduced. The device combines the advantages of the soft and rigid systems to deliver deterministic load compensation for the knee joint via the use of a semi-rigid chain structure for the tendon routing while effectively addressing any potential misalignment thanks to its flexible connection with the human body.  A novel strap routing arrangement is considered to distribute the tension load to the insensitive area of the human body. Validation experiments successfully demonstrated the working principle showing that the Exo-Muscle can effectively generate a desired assistive torque profile around the human knee joint. Future activity will focus on the development of a foot-wearable ground reaction force measurement device to permit the full mobility of the user while providing the measurement of ground reaction force, which is necessary for the knee torque estimation and the control. The effect of the device assistance on the user's metabolic cost and EMG activity of the muscles associated with the knee joint will also be studied.

\bibliographystyle{./IEEEtran}
\bibliography{./IEEEabrv,./RAL}

\end{document}